\documentclass[letterpaper, 10 pt, conference]{ieeeconf}  

\IEEEoverridecommandlockouts                              

\usepackage{graphicx}
\graphicspath{ {./images/} }
\usepackage{caption}
\usepackage{subcaption}
\usepackage{amsmath}
\usepackage{amsfonts} 
\usepackage[svgnames]{xcolor}
\usepackage[hyperfootnotes=false,colorlinks]{hyperref}\usepackage{lipsum}
\usepackage{multicol}
\usepackage{graphicx}
\usepackage{soul}
\usepackage{cite}
\usepackage{array}
\usepackage{multirow}
\usepackage{makecell}

\usepackage[export]{adjustbox}
\title{\LARGE \bf 
Robotic in-hand manipulation with relaxed optimization}

\author{Ali Hammoud$^{1}$, Valerio Belcamino$^{2}$, Quentin Huet$^{1}$, Alessandro Carfi$^{2}$, Mahdi Khoramshahi$^{1}$,\\
Veronique Perdereau$^{1}$ and Fulvio Mastrogiovanni$^{2}$
\thanks{This work was supported by the European project Index}
\thanks{$^{1}$A. Hammoud and V. Perdereau are with Sorbonne Universite, Institut des Systemes Intelligents et de Robotique,
ISIR, F-75005 Paris, France
        {\tt\small ali.hammoud, quentin.huet, mahdi.khoramshahi, veronique.perdereau@sorbonne-universite.fr}}%
\thanks{$^{2}$V. Belcamino, A. Carfì and F. Mastrogiovanni are with TheEngineRoom,  Department of Informatics, Bioengineering, Robotics, and Systems Engineering, University of Genoa, Via Opera Pia 13, 16145, Genoa, Italy
        {\tt\small valerio.belcamino@edu.unige.it, alessandro.carfi@dibris.unige.it, fulvio.mastrogiovanni@unige.it.}}
\thanks{© 2024 IEEE.  Personal use of this material is permitted.  Permission from IEEE must be obtained for all other uses, in any current or future media, including reprinting/republishing this material for advertising or promotional purposes, creating new collective works, for resale or redistribution to servers or lists, or reuse of any copyrighted component of this work in other works.}}

\begin{document}
\maketitle
\thispagestyle{empty}
\pagestyle{empty}
\begin{abstract}
Dexterous in-hand manipulation is a unique and valuable human skill requiring sophisticated sensorimotor interaction with the environment while respecting stability constraints. Satisfying these constraints with generated motions is essential for a robotic platform to achieve reliable in-hand manipulation skills. Explicitly modelling these constraints can be challenging, but they can be implicitly modelled and learned through experience or human demonstrations.
We propose a learning and control approach based on dictionaries of motion primitives generated from human demonstrations. To achieve this, we defined an optimization process that combines motion primitives to generate robot fingertip trajectories for moving an object from an initial to a desired final pose.
Based on our experiments, our approach allows a robotic hand to handle objects like humans, adhering to stability constraints without requiring explicit formalization. In other words, the proposed motion primitive dictionaries learn and implicitly embed the constraints crucial to the in-hand manipulation task.
\end{abstract}

\section{INTRODUCTION}
\begin{figure}[t]
  \centering
 \includegraphics[trim={0cm 5cm 0cm 0cm},clip, width=0.4\textwidth]{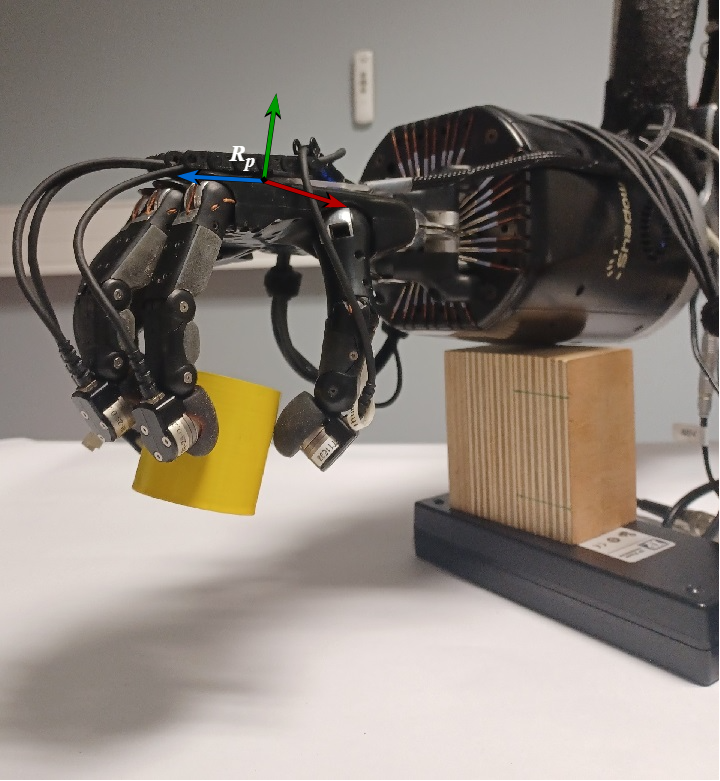}
  \caption{The Shadow Hand manipulating a cylinder. The reference frame for the palm ($R_{P}$) is shown, with the $x$, $y$, and $z$ axes in orange, green, and blue arrows, respectively.}
  \label{fig:example}
\end{figure}

Humans' ability to manipulate objects with dexterity is essential for interacting with their surroundings, allowing them to grasp, explore, and reorient objects. These skills are developed through a lifelong learning process that involves observing other people's behaviour and personal attempts and failures. %
A primary goal of human-robot interaction is to integrate robots into human-centred environments. However, the effectiveness of this integration depends on the robot's ability to move and operate in a human-like manner\cite{carfi2021hand}. Therefore, robots should be trained to manipulate unfamiliar objects and apply their prior knowledge to new situations (as displayed in Fig.~\ref{fig:example}). Additionally, robots should be able to learn new manipulation techniques by observing the actions of other agents. A robotic platform can achieve this by incorporating advanced perception tools and adaptable learning techniques to generate new smooth and dexterous manipulation operations.%
Planning a dexterous manipulation requires defining appropriate finger trajectories to reach a predetermined target configuration. Robotic manipulation planning strategies can be categorized into two groups: data-driven and analytical \cite{theodorou2010generalized, monahan1982state, prieur2012modeling, sundaralingam2017relaxed, sundaralingam2018geometric, fan2017real, 9659421}. In data-driven approaches, dexterous manipulation models are trained by robot trial and error or by observing human demonstrations \cite{theodorou2010generalized, monahan1982state, prieur2012modeling}. On the other hand, analytical solutions are based on robotics principles; complex tasks are divided into sets of elementary actions that solve sections of the more significant challenge \cite{sundaralingam2017relaxed,sundaralingam2018geometric, fan2017real}.

Most data-driven in-hand manipulation path planning approaches rely on Dynamic Movement Primitives (DMP) \cite{theodorou2010generalized}. By modifying a simple linear dynamical system with a non-linear component, DMP enables the creation of smooth movements of any shape \cite{ijspeert2002learning}. Therefore, DMP can learn from humans when data from human demonstrations are used to identify the non-linear components of a DMP system.%
On the other hand, analytical solutions for in-hand manipulation planning typically require modelling the robotic hand and its surroundings. Robot hand actions are decomposed into atomic processes, such as in-grasp reorientation and finger reallocation. In-grasp reorientation involves moving the object relative to the palm without changing the contact points between the fingers and the surface, for example, turning a dial. Sundaralingam and Hermans (2017) proposed an efficient solution to this problem with purely kinematic trajectory optimization \cite{sundaralingam2017relaxed}. In finger reallocation, the robot moves a single finger to a new contact location on the object while the remaining fingers maintain a stable grasp. Sundaralingam and Hermans (2018) \cite{sundaralingam2018geometric} and Fan et al. (2017)\cite{fan2017real} both used geometric approaches to offer solutions to finger gaiting. Classical methods provide predictable planning outcomes because the designer sets the constraints. However, fully modelling manipulation restrictions is complex, so these solutions can only approach atomic operations rather than the entire in-hand manipulation task. %
Both analytical and data-driven approaches have their limitations. Data-driven approaches require large training datasets and long training times. In contrast, the analytical approaches must model complex constraints to generate an entire in-hand manipulation trajectory. 

In our previous work\cite{9659421}, we introduced a novel approach that aims to balance the strengths and weaknesses of data-driven methods by extracting human skills. This is achieved through the creation of a dictionary of motion primitives, which are derived from human demonstrations. This dictionary comprises sparse vectors representing manipulation primitives and can be combined to generate human-like in-hand manipulation trajectories. In our subsequent research \cite{9900858}, addressing the generation of trajectories emerged as a pivotal concern. We leveraged a dictionary as the foundation for a more flexible path-planning algorithm. Our path planner successfully identified the necessary motion primitives to transition the hand from configuration A to configuration B while adhering to constraints such as finger reachable space, finger collisions, and the number of contact points with the object. Although these constraints were not explicitly imposed during the optimization process, they were implicitly extracted from human demonstrations. While this approach achieved human-like fingertip trajectories respecting environmental constraints, it did so without explicitly considering the object's pose. However, effective and realistic robotic manipulation requires considering the object's dynamics and kinematics.

In this work, we address the limitations of our previous solution by adopting a human in-hand manipulation solution that considers the relationship between the object pose and the fingertip positions. This solution can generate consistent fingertip trajectories given a desired change in the object pose. To accomplish this result, the data from human demonstrations must include the manipulated object's full pose (position and orientation) and fingertip positions. By integrating this comprehensive data, we can harness the learned primitives to plan in-hand manipulations that adhere to stability constraints while effecting the desired transformations on the handled object. Additionally, our proposed approach tackles the computational problem often associated with data-driven approaches by building a primitive dictionary from a small sample of demonstrations.

The rest of this paper is structured as follows. Section~\ref{sec:formalization} discusses the concept of a primitives dictionary and how it relates to in-hand manipulation. Section~\ref{sec:optimization} explains path planning using an in-hand manipulation dictionary. Section~\ref{sec:implementation} outlines the actions needed to incorporate an in-hand manipulation dictionary into the path-planning problem, including data collection, model training, and model testing. In Section~\ref{sec:exp}, we analyze robot simulation and experimental results. Finally, the last section presents discussions, conclusions and future work.

\section{Primitives Dictionary}
\label{sec:formalization}
Dictionaries of primitives are commonly used tools in literature for clustering and reducing the dimensionality of datasets. This method has broad applications, including computer vision \cite{cao2020nmf}, document clustering \cite{akata2011non}, astronomy \cite{berne2007analysis}, and motion planning \cite{vollmer2014sparse}. When focusing on human actions, the dictionary is considered to be composed of \textit{motion} primitives. Although there is no universal definition for motion primitives, they are often described as simple motions whose concatenation leads to complex human actions \cite{9412363}\cite{Flanagan2006ControlSI}. Furthermore, primitives' dictionaries can be learned from training data using the non-negative matrix factorization method (NMF) \cite{sra2005generalized}.

Let us begin with a set of demonstrations organized in the matrix form as $V \in \mathbb{R}^{n \times m}$, where $n$ and $m$ respectively represent the length and the number of demonstrations and each element of $V$ is $\geq 0$. Each column of $V$ denotes a single demonstration and we can apply the NMF algorithm that leads to the following decomposition:
\begin{equation}
    V=WH
\end{equation}
Where $W \in \mathbb{R}^{n \times l}$ is a matrix representing the extracted primitives and each column of $W$ represents a single primitive.
Given the ability of NMF to reduce the data dimensionality, the number of extracted primitives $l$ is considerably lower than the number of demonstrations $m$ (i.e. $l<m$).
Furthermore, $H \in \mathbb{R}^{l \times m}$ is also a matrix that represents the corresponding activation matrix and contains the weights to combine the primitives.
For instance, each $i$th column of $H$ encodes the weights that can be used to reconstruct the $i$th demonstration with a linear combination of the primitives. These two matrices $W$ and $H$ only contain non-negative elements

Using the extracted primitives, we can also generate new samples: given the desired behaviour to represent ($v$), it is possible to reconstruct it with a dictionary of primitives ($W$) by combining them with a set of weights ($h$):
\begin{equation}
   v = W h
    \label{eq:primitives}
\end{equation}
Each weight $h_{j}$ belonging to the activation vector $h$ influences to which extent every primitive $W_{j}$ contributes to the reconstruction of $v$.

\subsection{In-Hand Manipulation Primitives}
Object manipulation aims to achieve a specific object pose or trajectory through finger motions. Therefore, in-hand manipulations can be described as the evolution of fingertips’ positions and the object's pose over time, starting from an initial configuration and ending at the desired pose. Therefore, the problem can be decomposed into two main components: the trajectories of the fingers and the object's trajectory. With this assumption and recalling Eq.~\ref{eq:primitives}, we can approach the problem by describing the overall trajectory as:
\begin{equation}
    v = \left[ {\begin{array}{c}
    P(1)\\
    \vdots\\
    P(k)\\
    \vdots\\
    P(N) \\
  \end{array} } \right]\\
\end{equation}
where $v \in \mathbb{R}^{21N}$, $N$ is the number of time steps and 21, as we will see, is the number of features. In fact, each element $P(k)$ of $v$ represents the fingertips and object pose at time $k \Delta t$  and is defined as follows:
\begin{equation}
    P(k) = \left[ {\begin{array}{c}
    P_{1}(k \Delta t )\\
    P_{2}(k \Delta t )\\
    P_{3}(k \Delta t )\\
    P_{4}(k \Delta t )\\
    P_{5}(k \Delta t ) \\
    P_{O}(k \Delta t ) \\
  \end{array} } \right]\\
\end{equation}
\begin{equation*}
    \textit{s.t.} \hspace{0.1 cm} k \in [1,\dots,N]
\end{equation*}
with $P(k) \in \mathbb{R}^{21} $
and where
\begin{equation}
   \label{eq:finger}
   P_{i}(k \Delta t )  = \left[ {\begin{array}{c}
    x_{i}(k \Delta t )\\
    y_{i}(k \Delta t )\\
    z_{i}(k \Delta t )\\
  \end{array} } \right]\\
\end{equation}
and 
\begin{equation}
    \label{eq:object}
   P_{O}(k \Delta t )  = \left[ {\begin{array}{c}
    x_{O}(k \Delta t )\\
    y_{O}(k \Delta t )\\
    z_{O}(k \Delta t )\\
    \psi_{O}(k \Delta t )\\
    \theta_{O}(k \Delta t )\\
    \phi_{O}(k \Delta t )\\
  \end{array} } \right]\\
\end{equation}
 Eq.~\ref{eq:finger} and \ref{eq:object} represent the fingertip position in Cartesian space and the object pose, respectively. Therefore, there are 21 features at each time instant. Each of the five fingers is represented by x, y, and z coordinates, and the object pose includes its x, y, and z position as well as roll ($\psi$), pitch ($\theta$), and yaw ($\phi$).

Given these definitions, the problem of in-hand manipulation involves finding the fingertips and object trajectories from an initial pose $\hat P(1)$ to a final desired pose $\hat P(N)$.

This formalization leads to the definition of $W$ as follows:
\begin{equation}
    W = \left[ {\begin{array}{c}
    W(1)\\
    \vdots\\
    W(k)\\
    \vdots\\
    W(N) \\
  \end{array} } \right]\\
\end{equation}
where $W \in \mathbb{R}^{21N \times l}$ and $l$ is the number of primitives. Each element $W(k)\in \mathbb{R}^{21 \times l}$ represents the $l$ primitives at the $k$th of the 21 features describing the object and fingertips pose.
As a result of this formalization, we can write Eq.~\ref{eq:primitives} as below:
\begin{equation}
    \left[ {\begin{array}{c}
    P(1)\\
    \vdots\\
    P(N) \\
  \end{array} } \right]=
  \left[ {\begin{array}{c}
    W(1)\\
    \vdots\\
    W(N) \\
  \end{array} } \right]h
  \label{formal}
\end{equation}

We can see from this representation how each of the \textit{l} values of $h$ acts as a weight, assigning importance to each primitive of $W$ during the reconstruction of $v$.

\section{Generation of Manipulations}
\label{sec:optimization}
Once the dictionary of primitives has been created from the dataset of demonstrations, we can combine the primitives using weights to produce new in-hand manipulation trajectories. However, the created trajectories must respect the constraints of the robotic application. Before introducing the constraints and describing the process of generating in-hand manipulation, it is important to note that a few assumptions about the manipulation task limit the proposed approach:

\begin{itemize}
\item Only two factors can influence the pose of the object: the robot and gravity.
\item Both the robotic hand and the object are rigid.
\item The initial and final object grasps are stable.
\item Only the fingertips make contact with the object.
\end{itemize}
Now that the problem has been formalized and the initial assumption has been established, we can introduce the process of determining the proper weights to solve a particular manipulation. This is achieved through an optimization process that aims to minimize the following cost function:
\begin{equation}
   h=\arg\underset{h}{\min} \lVert P(1)- \hat P(1)\rVert^{2}+ \lambda \lVert P(N) - \hat P(N)\rVert^{2}
   \label{cost function}
\end{equation}
and by substituting Eq.~\ref{formal} into Eq.~\ref{cost function}, we get:
\begin{equation}
   h=\arg\underset{h}{\min} \lVert W(1)h- \hat P(1)\rVert^{2}+ \lambda \lVert W(N)h - \hat P(N)\rVert^{2}
   \label{cost function1}
\end{equation}
where the first and the second terms minimize the difference between the desired and achieved fingertips positions and object pose at steps $1$ and $N$, respectively. Lambda $ \lambda$  is the scalar weight fine-tuning the trade-off between the two cost components.

The optimization process must consider other constraints linked to the robotic hand kinematics while solving the problem. The i-th fingertip should be in the reachable workspace $\mathbb P_{i}$. That is mathematically defined as the set of points in three-dimensional space that the fingertip can reach. This set is typically constrained by the physical limitations of the robotic arm or hand controlling the fingertip.
\begin{equation}
    P_{i}(k \Delta t) \in \mathbb P_{i}, \forall i \in [1,5], \forall k \in [1,N]
    \label{eq:reachability}
\end{equation}
Additionally, each fingertip must comply with kinematic constraints on its instantaneous velocity.
\begin{equation}
    \dot{P}_{i,min} \leq \dot{P}_{i}(k \Delta t) \leq \dot{P}_{i,max}, \forall i \in [1,5], \forall k \in [1,N]
    \label{eq:velocity}
\end{equation}
Other constraints must also be considered. For example, the fingertips should not overlap during the motion. Additionally, during manipulation, the object must maintain contact with at least two fingertips of the Shadow Hand to ensure a stable grip, leveraging the compliant nature of the fingertips. This soft compliance allows for slight deformation upon contact, enhancing grasping capabilities, especially for irregular objects. Optimal stability is achieved by distributing forces through two fingertips, creating a moment about the object's centre of mass to prevent slippage or rotation. While any two fingers can establish contact, stability is maximized when forces oppose or act along different axes, simplifying grasp planning while maintaining effective manipulation \cite{Prattichizzo2008}.
We can express the first constraint by checking the Euclidean distances between fingertips for all the manipulation time:
\begin{equation}
 \min_{\forall j \in (i,5]} \lVert P_{i}(k \Delta t ) - P_{j}(k \Delta t ) \rVert \neq 0,
 \label{eq:collisions}
\end{equation}
\begin{equation*}
     \forall i \in [1,4], \forall k \in [1,N]
\end{equation*}
The constraint on the points of contact is represented by the distance between the fingertip ($P_{i}(k \Delta t )$) and the object surface, represented as a point-cloud ($O(k \Delta t)$) of 3D points ($o_{c}(k \Delta t)$, with $c \in [1,|O|]$). This representation presupposes a process for computing the point cloud influenced by the object's shape, dimension, and spatial pose. In light of this representation, the distance ($d_i(k \Delta t)$) between the $i$th fingertip and the object is considered as the distance between the fingertip and the closest point of the object point cloud:
\begin{equation*}
    d_i(k \Delta t) = \min_{\forall c \in [1,|O|]}\lVert P_{i}(k \Delta t) - o_{c}(k \Delta t)) \rVert
\end{equation*}
and the subset of fingertips in contact with the object ($P_c(k \Delta t)$)
\begin{equation}
    P_c(k \Delta t) = \{ P_{j}(k \Delta t) \hspace{0.1cm} | \hspace{0.1cm} d_j(k \Delta t) \leq \tau\}
    \label{eq:single-contact}
\end{equation}
where $\tau$ is a threshold that defines the maximum distance between a fingertip and an object to be considered in contact. Based on this definition, at each time step $k$, the cardinality of this subset must always be greater than 2 for soft fingertips to hold the object during manipulation.
\begin{equation}
    |P_c(k \Delta t)| \geq 2, \hspace{0.1 cm} \forall k \in [1,N]
    \label{eq:contacts}
\end{equation}

\subsection{Relaxed Optimization Problem}
Humans naturally adhere to the constraints presented in the previous section. Since our approach uses human data to train the motion primitives dictionary, we hypothesize that trajectories generated from the human manipulation dictionary will follow these constraints without explicitly including them in the optimization problem. This hypothesis applies to all constraints, except for the fingertips' velocity, because of the differences in velocity ranges between humans and robots. Therefore, the optimization process should find the weights $h$ that minimize Eq.~\ref{cost function1} while considering the fingertips' velocity constraint expressed in Eq.~\ref{eq:velocity}. The optimization process does not take into account the other constraints (Eq.~\ref{eq:reachability},\ref{eq:collisions},\ref{eq:contacts}). However, the generated trajectories will be evaluated to determine if they respect them.

\section{Implementation}
\label{sec:implementation}
Based on the problem statement introduced in Section~\ref{sec:formalization} and the optimization procedure detailed in Section~\ref{sec:optimization}, human demonstrations of in-hand manipulations are required to generate in-hand manipulation. These demonstrations should include both the 3D positions of the fingertips and the object pose.

\subsection{Data Acquisition}
\begin{figure}[t]
  \centering
  \includegraphics[width=0.35\textwidth]{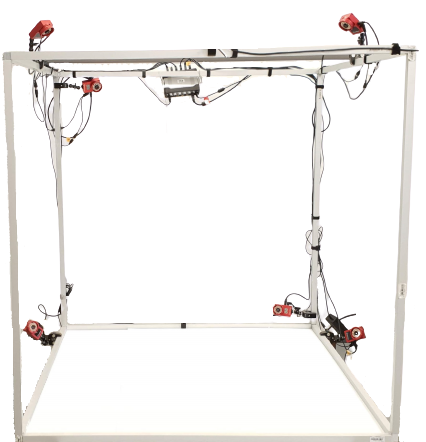}
  \caption{Six Flex-3 cameras arranged on a one-meter cube frame.}
  \label{fig:setup}
\end{figure}

\begin{figure}[t]
  \vspace{0.2cm}
    \centering
    \begin{subfigure}{0.45\columnwidth}
        \centering
        \includegraphics[width=\columnwidth]{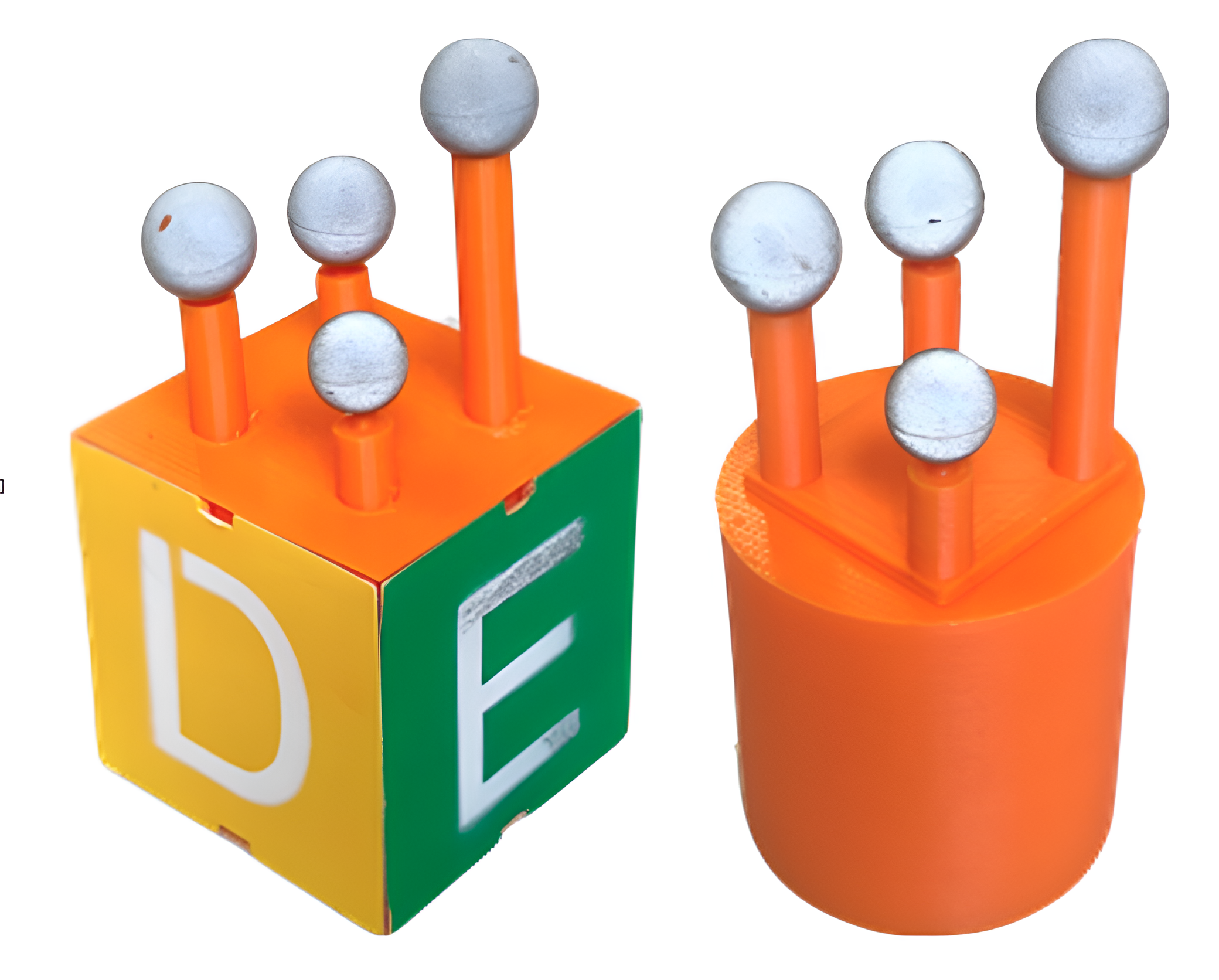}
    \end{subfigure}\hfill
    \begin{subfigure}{0.48 \columnwidth}
        \centering
        \includegraphics[width=\columnwidth]{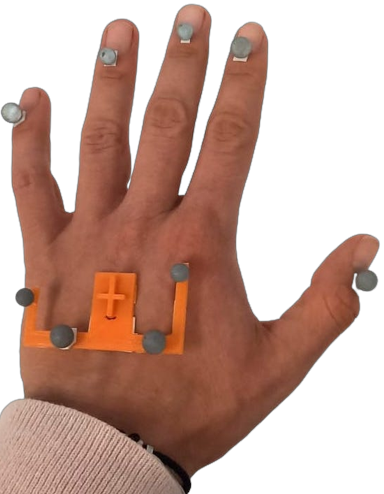}
    \end{subfigure}
    \caption{On the left, the markers are placed on two different objects, while on the right, the markers are placed on a hand. The back of the hand and the two objects have multiple markers on rigid supports for tracking their orientation.}
    \label{fig:markers}
\end{figure}

The acquisition of human demonstrations has been performed using a motion capture (MoCap) system comprising six OptiTrack Flex-3 cameras\footnote{\scriptsize\url{https://optitrack.com/cameras/flex-3/}} (see Fig.~\ref{fig:setup}). These infrared cameras can track the position of small reflective objects within the field of view of at least three cameras (see Fig.~\ref{fig:markers}). The same concept can be extended to determine orientation, but, in this case, it is necessary to define a rigid body by attaching three or more markers to the object's surface. The Motive\footnote{\scriptsize\url{https://www.optitrack.com/software/motive/}} software acquires the information mentioned above at a frequency of 100 Hz. In case of interruption in the tracking, the user can fill in the missing parts of the sequences through cubic interpolation.

We positioned markers to record the human demonstrations as shown in Fig.~\ref{fig:markers}. We used one reflective sphere for each finger on the distal phalanges and a 4-marker rigid body positioned on the hand back to track the palm pose.

For the experiments, we considered two differently shaped objects: a cube with a 5-centimeter edge and a cylinder with a diameter and height of 5 cm. We used the same 4-marker rigid body configuration to track the objects, as shown in Fig.~\ref{fig:markers}, ensuring that the markers' centroid corresponded to their axis of symmetry. This choice allowed us to determine the position of the centre of the objects by applying a simple translation in the local system of reference.

Due to the limited number of cameras and the close positioning of the markers during manipulation, the markers could easily occlude each other, resulting in poor tracking results. To solve this issue, we restricted the manipulations to a specific hand pose in space, i.e., the palm facing up in the middle of the tracking area, and disallowed any significant wrist rotation. These precautions allowed us to achieve high measurement precision by minimizing occlusions and collisions. 

We conducted six 5-minute trials for each object using the described setup. Each trial included rotations and translations along multiple axes and required the object's entire weight to be held exclusively by the fingertips. As previously mentioned, to ensure that the manipulations were solely resulting from finger motions and to optimize the recordings' quality, wrist motions were not allowed. In addition, the rotations of the objects were limited by the markers mounted on them and the need to avoid occlusion.

The recording phase resulted in two datasets of in-hand manipulations, one for each object shape. Each of the two datasets contains 36 minutes of recordings split into a 30-minute training set and a 6-minute test set.

\subsection{Model Training and Testing}
Before the training process can begin, the collected data must go through three processing phases. At first, all the data is transformed into the hand-palm reference. Based on the first assumption presented in Section~\ref{sec:optimization}, we will only focus on the manipulation components related to the fingertips. Following this assumption, we should consider the hand palm static and only focus on the motion of the fingertips and objects. In the second phase, the data is filtered by a low pass filter with a cut-off frequency of 20Hz. The cut-off frequency is based on a study by Xiong and Quek (2006) \cite{xiong2006hand}, which indicates that a 10Hz sampling rate is sufficient to detect human hand motion. The third and final step ensures that the dataset $V$ contains only positive values. This is ensured by offsetting all position data by $0.8$ \textit{m} and all object orientation data by $2\pi$.

After the processing phase, we followed the training and segmentation steps as presented in our previous research \cite{9659421,9900858}. The processed data was segmented into sequences of 1 second each and stacked to build the columns of the $V$ matrix. We then applied the non-negative matrix factorization method (NMF) to obtain the $W$ dictionary from the $V$ data matrix. This procedure was performed separately for the cube and the cylinder, resulting in two distinct dictionaries of primitives. The reason for using 1-second segments is to model intermediate configurations of the in-hand manipulation and simplify the optimization process. The final dictionaries each contain 200 motion primitives. The processing and training phase took 50 minutes using a single graphical process unit.

Finally, we tested the dictionary's ability to recreate human in-hand manipulation trajectories\cite{9659421}. We collected errors between the original and recreated trajectories (see Table~\ref{error_table}) by considering both the Euclidean distances of the positions and the difference in object orientation. Both dictionaries showed good accuracy, with a mean error on fingertips position equal to $0.6$ \textit{mm} and a standard deviation of $0.45$ \textit{mm}. Additionally, the object orientation and position error had similar results, with mean values of $0$ \textit{rad} and $0.4$ \textit{mm}, and standard deviations of $0.55$ \textit{rad} and $0.3$ \textit{mm}, respectively.
{\def\arraystretch{1.2}
\begin{table}[t]
\vspace{0.2cm}
    \centering
    \caption{This table shows Euclidean distances between the real and recreated fingers/objects positions, along with object-orientation differences.}
    \begin{tabular}{ l | c | c }
    
    \multicolumn{1}{l}{}   & \multicolumn{1}{c}{Cube Dictionary} & \multicolumn{1}{c}{Cylinder Dictionary}\\
    \hline
    \multicolumn{3}{l}{\textbf{Fingers}} \\
    \hline
  Thumb & \; $0.4755 \pm 0.4470$ \textit{(mm)} & $0.6912 \pm 0.6903$ \textit{(mm)}\\ 
  \hline
  Index  & \; $0.4402 \pm 0.4246$ \textit{(mm)}  & $0.7874 \pm 1.0882$ \textit{(mm)}\\ 
  \hline
  Middle  & \; $ 0.4276 \pm 0.3915$ \textit{(mm)} & $0.6846 \pm 0.8708$ \textit{(mm)}\\ 
  \hline
  Ring  & \; $0.4462 \pm 0.4398$ \textit{(mm)} & $0.5591 \pm 0.7200 \textit{(mm)}$  \\ 
  \hline
  Little & \; $0.4493 \pm 0.4271$ \textit{(mm)} & $0.5511 \pm 0.7042 \textit{(mm)}$ \\ 
  \hline
  \multicolumn{3}{l}{\textbf{Object}} \\
  \hline
  Translation & \; $0.5108 \pm 0.5106$ \textit{(mm)} & $0.8176 \pm 1.0774$ \textit{(mm)} \\ 
  \hline
  Roll & \; $0\pm 0.0132$ \textit{(rad)}& $0.0001 \pm 0.0179$ \textit{(rad)} \\ 
  \hline
  Pitch & \; $0.0001 \pm 0.0198$ \textit{(rad)} & $0 \pm 0.0175$ \textit{(rad)} \\ 
  \hline
  Yaw & \; $0 \pm 0.0017$ \textit{(rad)} & $0 \pm 0.0039$ \textit{(rad)} \\
  \hline
    \end{tabular}
    \label{error_table}
\end{table}%
}

After confirming that the two dictionaries can accurately recreate human motion, the following step is integrating them into a pipeline for generating in-hand manipulation for a robotic platform. To solve the optimization problem introduced in Section~\ref{sec:optimization}, we used IBM ILOG's CPLEX optimization studio in C++. This software enabled us to tackle complex optimization tasks based on linear and mixed linear programming. The optimizer requires initial and final desired stable grasp configurations to obtain fingertip trajectories in the Cartesian space. Next, an Inverse Kinematics algorithm, based on dynamics-based recursive linearization (RDBL) principles, converts the fingertip trajectories from Cartesian to joint space. Finally, the trajectory is used to control a Shadow Hand. We have created and tested various trajectories representing different in-hand manipulation scenarios. The experimental setup and results are discussed in the following section.

\section{Experiments and Results}
\label{sec:exp}
Our method aims to overcome the limitations of analytical and data-driven in-hand manipulation approaches. We achieve this by generating a non-complex method that can cover in-hand manipulation and finger gaiting without requiring a heavy training process or complex representation of the constraints.

We tested the ability of our system to achieve the desired object pose by measuring its position and rotation after each trajectory. Additionally, we compared the performance of our method to analytical and data-driven approaches.

\subsection{System overview}
\label{system_overview}
\begin{figure}[t] 
\vspace{0.2cm}
 \centering
 \includegraphics[trim={0 0 0 1cm},clip,width=0.4\textwidth]{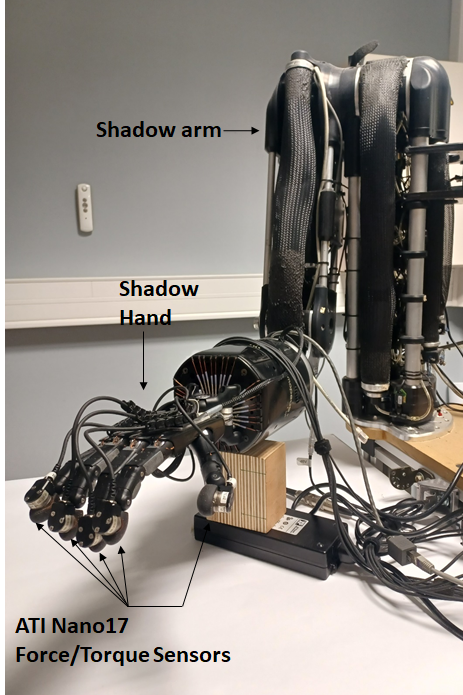}
 \caption{The picture shows the platform used for the experiments. This includes the Shadow Hand, a mechanical arm, and 5 force sensors placed on the fingertips (AT Nano17).}
 \label{platform_titled}
\end{figure}

We implemented the proposed approach for generating in-hand manipulations using the ROS middleware, and we tested it on an anthropomorphic robotic hand (refer to Fig.~\ref{platform_titled}). The experimental setup consists of a Shadow Hand\footnote{\scriptsize\url{https://www.shadowrobot.com/dexterous-hand-series/}} connected to a Shadow Arm and custom-designed fingertips with 6-axis ATI nano17 force and torque sensors. A state-of-the-art algorithm processes data from the fingertips’ sensors to estimate the contact of the fingers with the object based on force and torque sensor measurement \cite{liu2012surface}. Additionally, the setup comprised a goniometer and a calliper for measuring the orientations and translations of the objects. 

\subsection{Results}
The training approach is based on extracting manipulation features unique to each object. Therefore, for each object, we used the corresponding dictionary.

Since we only considered in-hand manipulations, we fixed the Shadow Hand, adding stable support, and all object trajectories were executed through finger actuation. Our experimental scenario was designed to consider the most common actions among those described by Elliott et al. (1984)\cite{elliott1984classification} in their classification of in-hand manipulations. Thus, the three operations we considered were rotation on the $x$ and $y$ axes, and translation on the $y$ axis. These axes correspond to the palm reference system $R_{P}$ in Fig.~\ref{fig:example}.


After completing each trajectory, we measured the object's orientation using a goniometer and its translation with a calliper.
The performance of our algorithm was tested on 21 in-hand manipulation trajectories, evenly distributed as follows: seven rotations around the $x$ axis, seven rotations around the $y$ axis, and seven translations along the $y$ axis.
For rotation actions, the objects were rotated within an angle range of 15 to 20 degrees, while translation actions involved displacements of 5 to 10 cm.
Table~\ref{tab:my_label} shows the errors associated with each movement, indicating their average, minimum, and maximum values. 

\begin{figure*}[t]
\vspace{0.1cm}
 \centering
 \includegraphics[width=1\textwidth, angle=0]{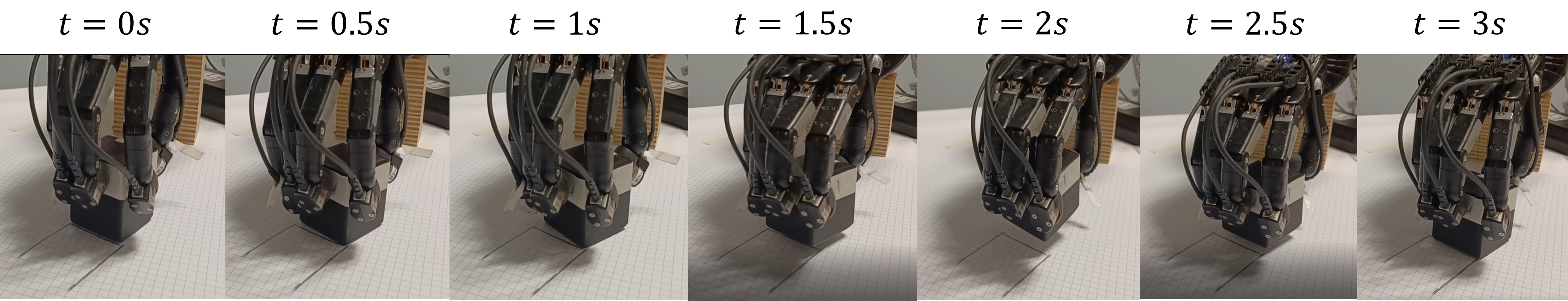}
 \caption{This series of images shows the manipulation of a cube over time. First, the Shadow Hand rotates the object around the y-axis. Then, the Shadow Hand translates the object along the y-axis. Finally, the Shadow Hand performs a composite motion to return the object to its initial pose. (The system of reference $R_{P}$ is the same considered in Fig.~\ref{fig:example}).}
 \label{snapshot}
\end{figure*}

For the first constraint, we defined the reachable workspace $\mathbb{P}_i$ for each finger as the entire workspace of every fingertip. Next, we checked if the fingertip positions satisfied the reachability constraint described in Eq.~\ref{eq:reachability}. All the points generated with our approach met the reachability constraint.

For the finger collision constraint, we computed the minimum finger-to-finger distance as defined in Eq.~\ref{eq:collisions}. We detected five times instants with a collision for the cube and six times instants for the cylinder over the twenty-one trajectories generated for each object. Therefore instances in which the collision constraint is not satisfied are rare and, in any case, it never caused a task failure.

Finally, we recorded the contact data between hand fingertips and the object using the fingertips sensor addressed in subsection \ref{system_overview} to satisfy the minimum contact points between the fingertips and the object constraint. For all generated trajectories, a minimum of two fingers were in contact with the object at any given time. This is sufficient since the fingertips of the Shadow Hand provide soft contact.

After testing the ability of the relaxed optimization solver to generate trajectories that adhere to in-hand manipulation constraints, we evaluated our method for generating complex tasks. To achieve this, we created trajectories resulting from a composite of manipulations. First, the Shadow Hand rotated the object, then translated it, and finally performed a complex motion in which the object returned to its initial pose, involving a composite rotation and translation. As shown in Fig.~\ref{snapshot}, the Shadow Hand successfully returned the object to its initial pose after the series of three actions. A video demonstrating the entire procedure is available\footnote{\url{https://www.youtube.com/watch?v=2uszlWyuYWw}}.

Our approach to the in-hand manipulation task focuses on re-grasping and finger gaiting from a high-level view. We aim to verify whether the path planner can execute both re-grasping and finger gaiting actions. By analyzing recorded data on how the fingertips contact the object, we demonstrate that the number of contacting fingers varies over time, which confirms finger gaiting. Additionally, if finger gaiting is detected, any observed changes in the object's orientation can be considered strong evidence that re-grasping has occurred.

\begin{figure*}[t]
    \hspace{0.2cm}
    \centering
    \begin{subfigure}{0.4 \textwidth}
        \centering
        \includegraphics[height = 0.8\textwidth, right]{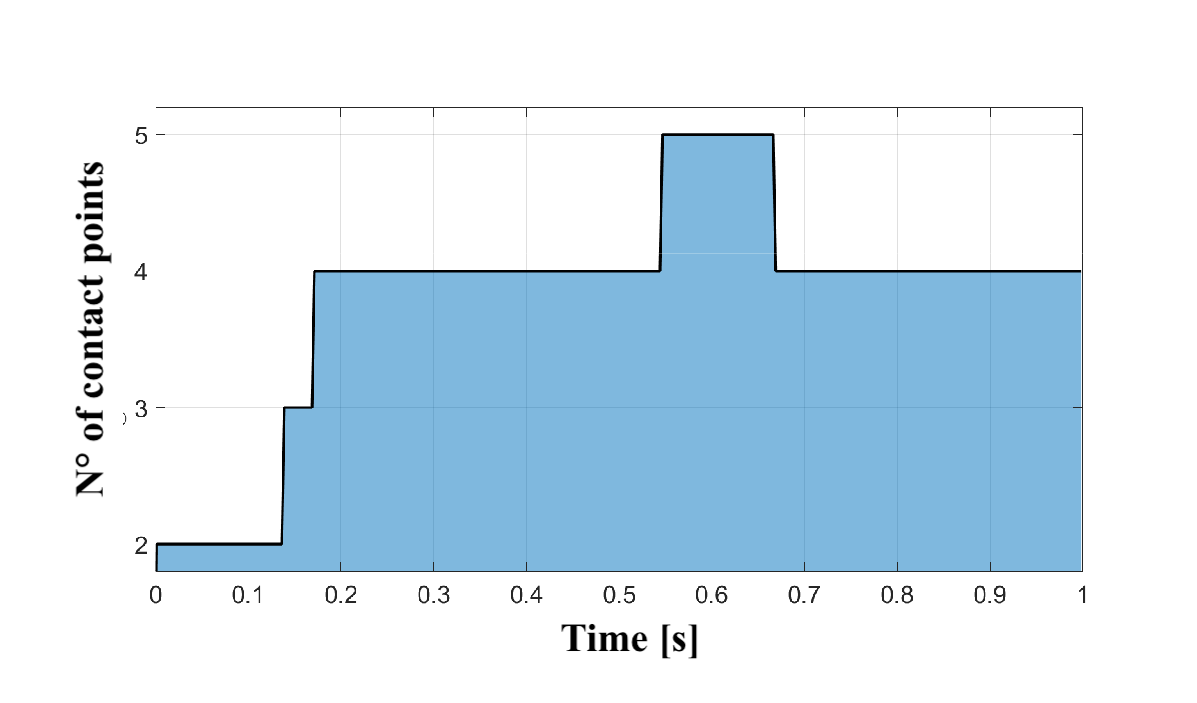}
    \end{subfigure}
    \begin{subfigure}{0.4 \textwidth}
        \centering
        \includegraphics[height = 0.8\textwidth]{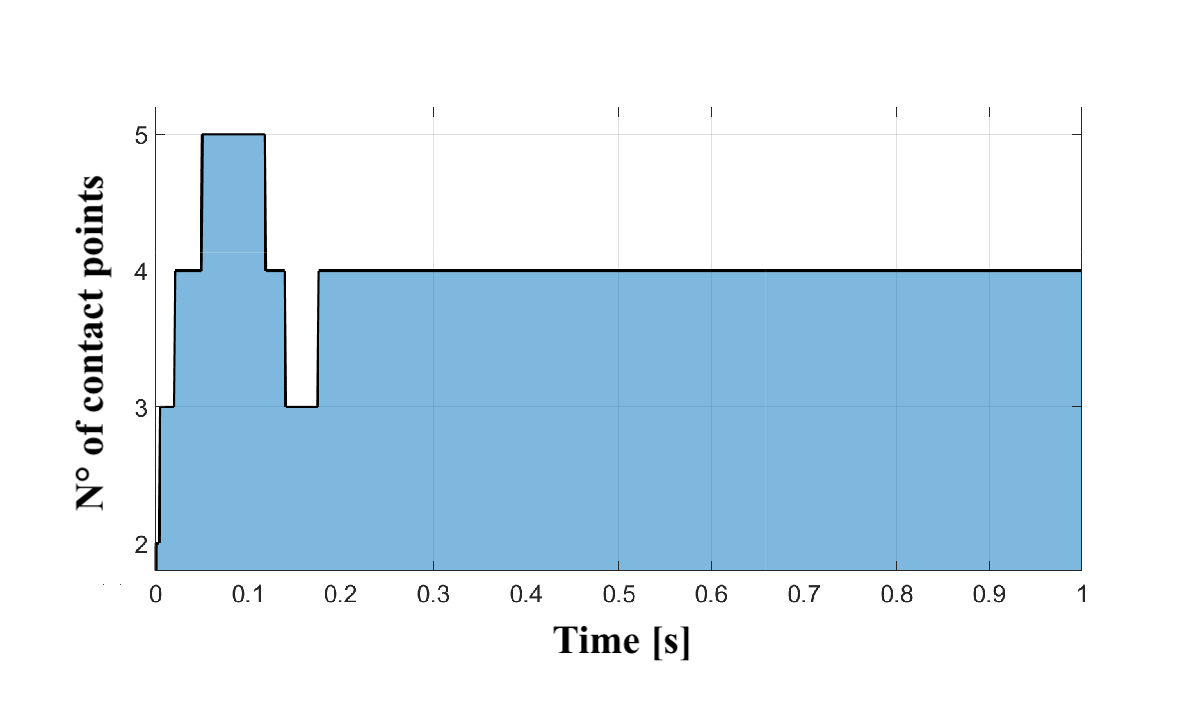}
    \end{subfigure}
    \caption{The picture illustrates the number of fingers in contact with a cube while executing two manipulations. The left and right graphs display respectively data for pure rotation and translation. The over-time variation in the number of contact points demonstrates the method's ability to generate finger gaiting.}
    \label{fing_contact}
\end{figure*}

In Fig.~\ref{fing_contact}, we present two examples of the variation in the number of contacts between the object and hand during object rotation and translation. This means that the object is undergoing in-hand re-grasping since its pose is changing, and finger gaiting is occurring since the number of fingers in contact with the object changes during the action.

{\def\arraystretch{1.3}
\begin{table}[t]
\vspace{0.2cm}
    \centering
    \caption{Performance of the algorithm on rotation and translation motions over the different in-hand manipulation trials. The mean error and the full error range are presented.}
    \begin{tabular}{ l | c | c }
        \multicolumn{1}{l}{} & \multicolumn{1}{c}{Cube} & \multicolumn{1}{c}{Cylinder} \\
         \hline
        \multirow{2}{*}{Translation error on y-axis [\textit{mm}]} &  $-3.183$   & $0.280$  \\ 
         &  $[-9,-1]$ &  $[+6,-2]$\\ 
        \hline
        \multirow{2}{*}{Rotation error on x-axis [$^{\circ}$]} & $-0.573$ & $0.785$ \\ 
         & $[-1,+1]$ & $[-1.5,+2.5]$ \\
        \hline
        \multirow{2}{*}{Rotation error on y-axis [$^{\circ}$]} & $-1.089$ & $-0.499$ \\ 
         & $[-2,-1]$ &  $[-2,+3]$ \\ 
        \hline
    \end{tabular}
    \label{tab:my_label}
\end{table}%
}

The median and maximum time for generating a trajectory in Cartesian space and applying inverse kinematics to find it in joint space is $150$ and $350$ \textit{ms}, respectively. We compared our results for the same problem with a classical approach, as presented in \cite{sundaralingam2018geometric}, which represents manipulations as a sequence of finger gating and in-hand re-grasping. In this approach, an optimization process finds the optimal trajectory for each atomic action. We tested two different types of solvers for the optimization problem. The first solver aims to reduce the signed distance between the reachable workspace and the desired contact points, and it showed a median and maximum planning time of $729.05$ and $3513.96$ \textit{s}, respectively. The second solver is based on the singular value decomposition, and it showed a median and maximum planning time of $75$ and $134.275$ \textit{s}. Therefore, our method can create a trajectory including in-hand re-grasping and gating faster than this analytical solution alternating between the two atomic actions.

Finally, it is important to evaluate whether the fast response in trajectory generation comes at the expense of lengthy procedures for data acquisition and dictionary training. In our study, collecting human demonstrations and conducting dictionary training took approximately 50 minutes using a single GPU. In contrast, using the data-driven strategy described in \cite{andrychowicz2020learning} takes up to 50 hours of training on the experimental set-up using 8 GPUs, roughly equivalent to 291.6 hours using a single GPU.

\section{Conclusions}
This paper presents a method for generating fingertip trajectories for in-hand manipulation using motion primitive dictionaries. The approach utilizes human demonstrations to extract a manipulation dictionary that implicitly respects in-hand manipulation constraints.

The proposed approach has been tested for in-hand manipulations of two objects (a cube and a cylinder) and could find trajectories to reach the desired pose while respecting three constraints that characterize in-hand manipulation (reachability, finger collision, and minimum contact points). These results show that the proposed approach retains the constraints from human demonstrations without requiring a formal representation. The generated trajectories solve the in-hand manipulation problem by including in-grasp re-orienting and finger gaiting as actions to move the object to the desired pose. Furthermore, our approach achieves competitive results in terms of trajectory generation time and training time compared to analytical and data-driven approaches, respectively.

In conclusion, we want to mention the possibility of adapting our method to robotic hands that do not replicate human anatomy. Our approach, centred on observing the poses of the fingertips and not relying on detailed information about the kinematic chains of the fingers, suggests that it could be extended to non-anthropomorphic robotic hands. However, this extension comes with certain assumptions and limitations. We presuppose that the robotic fingers possess the necessary dexterity for flexion-extension, adduction and abduction movements. Furthermore, our method implicitly assumes the use of a five-finger configuration. This is because the constraints we apply to ensure object stability are derived from a human motion dictionary, which naturally presupposes a hand with five fingers. While our method holds promise for broader application, these considerations highlight the need for careful adaptation when using robotic hands with configurations that deviate from the human one.

Given these results, this approach's trajectories can enable robotic hands to perform dexterous manipulations. However, future work should extend the proposed approach to more advanced manipulation tasks, implementing more precise control strategies and relaxing our initial assumptions.





\section*{ACKNOWLEDGMENT}
This work is supported by the CHIST-ERA (2014-2020) project InDex and received funding from Agence Nationale de la Recherche (ANR) under grant agreement No. ANR-18-CHR3-0004 and the Italian Ministry of Education and Research (MIUR).
\bibliography{mybib}{}
\bibliographystyle{ieeetr}

\end{document}